\documentclass[lettersize,journal]{IEEEtran}
\usepackage{amsmath,amsfonts}
\usepackage{array}
\usepackage[caption=false,font=normalsize,labelfont=sf,textfont=sf]{subfig}
\usepackage{textcomp}
\usepackage{stfloats}
\usepackage{url}
\usepackage{verbatim}
\usepackage{graphicx}
\usepackage{cite}
\usepackage{amsmath,amssymb,amsthm}
\usepackage{booktabs}
\usepackage{hyperref}
\usepackage{booktabs}
\usepackage{longtable}
\usepackage[table,xcdraw]{xcolor}
\usepackage{cite}
\usepackage{amsmath,amssymb,amsfonts}
\usepackage{graphicx}
\usepackage{textcomp}
\usepackage{xcolor}
\usepackage{algpseudocode}
\algrenewcommand\algorithmicrequire{\textbf{Input:}}
\algrenewcommand\algorithmicensure{\textbf{Output:}}
\usepackage{algorithm}
\usepackage{bm}
\usepackage{booktabs}
\usepackage{multirow}
\usepackage{makecell}
\usepackage{xcolor}
\usepackage{romannum}
\begin{document}
\pagenumbering{arabic}
	\setcounter{page}{1}
\title{EvoTSC: Evolving Feature Learning Models for Time Series Classification via Genetic Programming}

\author{Xuanhao Yang, Bing Xue, \IEEEmembership{Fellow, IEEE}, and Mengjie Zhang, \IEEEmembership{Fellow, IEEE}
	\thanks{This work has been submitted to the IEEE for possible publication. Copyright may be transferred without notice, after which this version may no longer be accessible.}
	\thanks{Xuanhao Yang, Bing Xue, and Mengjie Zhang are with the Centre for Data Science and Artificial Intelligence \& School of Engineering and Computer Science, Victoria University of Wellington, PO Box 600, Wellington 6140, New Zealand (e-mails: xuanhao.yang@vuw.ac.nz; bing.xue@ecs.vuw.ac.nz; and mengjie.zhang@ecs.vuw.ac.nz).}
}

\maketitle

\begin{abstract}
Time series classification is an important task across diverse domains. However, its practical application is often limited by scarce labeled data and the requirements for substantial computational resources. To address these challenges, this paper proposes EvoTSC, a novel genetic programming approach designed to automatically evolve lightweight feature learning models for time series classification. The core of EvoTSC is a multi-layer program structure that embeds diverse forms of prior knowledge into evolution, guiding the search toward operations effective for time series classification. A tailored Pareto tournament selection strategy is further proposed to favor models that perform consistently across varying folds of the training data, thereby promoting better generalization. Furthermore, an ensemble classifier called EvoTSC-E is built upon the evolved feature learning models for scenarios where computational resources are less constrained. This work focuses on equal-length univariate time series. Experiments show that the evolved feature learning models significantly outperform hand-crafted feature extraction baselines and most  time series foundation model baselines, while achieving more wins than losses against the self-supervised neural baselines. EvoTSC-E attains the best average rank among all 17 compared methods. Further analyses verify the contribution of each component and the resource efficiency of the evolved EvoTSC models.
\end{abstract}


\section{Introduction}
\label{s1}
Time series classification (TSC), the task of assigning categorical labels to temporal sequences, plays a critical role in diverse domains, including medical diagnosis \cite{lin2019medical}, industrial monitoring \cite{polge2020case}, and financial analysis \cite{liang2023improving}. Compared with images and text, time series data lack intuitive semantic clarity, making accurate annotation heavily reliant on costly domain expertise. Consequently, label scarcity is a common challenge in real-world TSC tasks \cite{LTSRL,TSPM}. To overcome this challenge, feature-based TSC approaches have attracted considerable attention \cite{FeaTSC}. By learning discriminative representations guided by explicit objective functions, these methods offer strong potential for facilitating downstream classification even under limited labeled data.

Early explorations in this area relied heavily on domain expertise to design and extract hand-crafted features \cite{catch22}. Such approaches typically summarize raw time series by computing a fixed set of attributes, such as statistical characteristics (e.g., median, maximum), shape descriptors (e.g., slope), and autocorrelation coefficients \cite{tsfel}. A notable strength of these methods lies in their high interpretability, as the extracted features can carry clear physical or statistical meaning \cite{hctsa}. However, relying on a fixed set of predefined features for datasets from various domains inherently limits model flexibility and adaptability.

Recent advancements in feature-based TSC methods have shifted from typical hand-crafted feature extraction toward neural network-based feature learning \cite{ts2vec,frera}. These approaches commonly pre-train a neural network (NN) with a self-supervised objective, preventing the learned features from being overly tailored to specific decision boundaries and thereby enhancing generalizability to the downstream classification task \cite{Kang2020Decoupling,cbd}. Due to the extensive parameter space, these methods can learn flexible, dataset-adaptive features, yielding enhanced classification performance. Following this trend, large foundation models with substantially greater parameter counts are increasingly being explored for TSC \cite{mantis,moment}. While the large scale of these models makes pre-training on the target dataset alone insufficient, it also enables them to leverage large-scale external data for pre-training, producing transferable features for the target task \cite{tivit}.

While neural network-based approaches, including foundation models, have achieved remarkable success in learning discriminative features, several fundamental limitations remain. First, the large number of parameters demands substantial computational resources for inference, posing practical challenges for deployment in resource-constrained environments, such as wearable health devices \cite{ECGTCN} and industrial monitoring sensors \cite{saeed2025deep}. Second, these approaches typically operate as black-box systems with limited interpretability, which is a significant concern in safety-critical applications \cite{interpretable}. These limitations collectively motivate the exploration of an alternative paradigm.

Genetic programming (GP) is an evolutionary computation technique inspired by the principles of natural selection \cite{banzhaf1998genetic,zhangtevc1}. By embedding domain-specific knowledge within carefully designed program structures, GP has been successfully applied to feature learning tasks across diverse domains \cite{wutevc,peng2}. Within these methods, a GP individual is an explicit composition of executable operations. Each learned feature can be traced back to a concrete sequence of transformations, making the resulting models potentially interpretable \cite{wutevc,bitevc}. Moreover, GP solutions are highly lightweight: once evolution finishes, prediction reduces to executing a compact model, eliminating the need for substantial computational resources and high inference latency \cite{11030860,srlinear}.

Although GP has strong potential to address these limitations, research on GP for feature learning specifically tailored to TSC remains very limited. To bridge this gap, this paper proposes EvoTSC, a GP-based method designed to evolve lightweight and effective feature learning models for TSC. The approach is built upon a novel multi-layer GP program structure that systematically organizes segment detection, domain transform, content-adaptive patching, feature extraction, and feature concatenation into a hierarchical search space. This structural design aims to embed diverse forms of prior expert knowledge directly into the evolutionary process. Meanwhile, to mitigate the common overfitting problem in TSC tasks, a tailored modification to the Pareto tournament selection \cite{paretotournament} is introduced, designed to promote the evolution of models that perform consistently well across varying folds of the training data. Additionally, motivated by the effectiveness of ensemble learning in TSC \cite{inceptiontime,hivecote} and by the fact that the models evolved by EvoTSC can be easily combined into an ensemble, we further develop EvoTSC-E\footnote{Throughout this paper, EvoTSC refers to the proposed GP method that evolves feature learning models, while EvoTSC-E refers to the ensemble classifier built upon the feature learning models evolved by EvoTSC.}, an ensemble classifier. Each ensemble member consists of an evolved feature learning model and its base classifier. EvoTSC-E thus trades additional computational cost for enhanced classification performance. Note that this work focuses on univariate time series of equal length. The main objectives are summarized as follows:
\begin{enumerate}
	\item Propose a novel multi-layer program structure specifically designed for feature learning in TSC, enabling the evolution of lightweight and effective feature learning models.
	
	\item Explore an improved selection strategy to mitigate the overfitting problem commonly encountered in TSC from an evolutionary perspective.
	
	\item Develop an ensemble classifier built upon the evolved feature learning models to further improve the classification performance.
	
	\item Evaluate the effectiveness of the evolved feature learning models and the classification performance of EvoTSC-E through comprehensive comparisons with representative TSC methods.
	
	\item Conduct further analyses to verify the contribution of each component of EvoTSC and evaluate the resource efficiency of the evolved feature learning models.
\end{enumerate}

\section{Related Work}
\label{s2}
\subsection{Feature-based Time Series Classification Approaches}
The primary objective of feature-based time series classification approaches is to transform a time series into discriminative feature vectors, thereby enhancing the performance of downstream classification tasks. This subsection reviews typical work of three different kinds of approaches.

\subsubsection{Hand-crafted Feature Extraction Approaches}
A foundational work in this area is hctsa \cite{hctsa}, which consolidates decades of interdisciplinary time series analysis methods into a large, interpretable feature library, computing over 7{,}700 features per input series. Building upon the hctsa library, catch22 \cite{catch22} addresses the computational overhead and feature redundancy of such large-scale feature sets by distilling a compact canonical subset. Another widely used approach is TSFresh \cite{tsfresh}, which offers an intermediate feature dimensionality between hctsa and catch22 by computing a pre-defined set of features spanning the temporal, statistical, and frequency domains. While computationally efficient and interpretable, hand-crafted features are inherently static and heavily reliant on domain expertise. Without data-driven optimization to adapt to dataset-specific distributions, these methods may yield irrelevant features.

\subsubsection{Neural Network-based Approaches}
To overcome the issues of hand-crafted features, recent methods employ neural networks pre-trained with self-supervised objectives to learn adaptive features directly from data. Among these, contrastive learning is a widely adopted paradigm, aiming to shape the latent space by pulling together semantically similar views while pushing apart dissimilar ones \cite{softclt}. For instance, Yue et al. \cite{ts2vec} propose a hierarchical contrastive framework that contrasts augmented context views across multiple temporal resolutions, enabling the model to capture contextual information at different granularities. From a frequency-domain perspective, Tian et al. \cite{frera} argue that frequency representations exhibit desirable properties for constructing semantic-preserving views and propose a frequency-refined augmentation strategy. This strategy separates and preserves important frequency components while adaptively modifying the remaining ones. Beyond contrastive learning, masked modeling methods represent another popular paradigm, which trains models to reconstruct missing content from visible context, thereby encouraging the learned features to preserve the underlying structure \cite{simmtm}. Cheng et al. \cite{timemae} propose a representative method that employs a decoupled architecture, consisting of a visible encoder that processes only the unmasked patches and a masked decoder that reconstructs the missing portions, facilitating effective feature learning. Han et al. \cite{cbd} address the issues of feature homogenization and spectral bias by introducing an auxiliary content-aware balanced decoder. This module adapts interactions to local content variations and explicitly rebalances spectral energy.

\subsubsection{Foundation Model-based Approaches}
Unlike the above NN-based methods, which train a model solely on the target dataset, foundation model approaches involve pre-training on vast external datasets to extract highly transferable features. A notable example is Mantis \cite{mantis}, a lightweight time series foundation model (TSFM) specialized for TSC tasks. Built upon a vision Transformer (ViT) backbone and pre-trained via contrastive learning on 1.89 million time series samples, Mantis achieves remarkable zero-shot classification performance. TiViT \cite{tivit} explores a cross-modality strategy by transforming one-dimensional time series into two-dimensional images. This allows the model to leverage the representational power of large-scale ViT pre-trained on billions of natural images. 

In summary, NN-based feature learning approaches, including foundation models, have achieved remarkable success in learning flexible and discriminative features. However, these models typically operate as black boxes \cite{interpretable} and demand substantial computational resources during inference. These limitations collectively motivate the exploration of alternative paradigms that offer more inspectability while remaining sufficiently lightweight for resource-constrained TSC applications.

\subsection{GP for Feature Learning}
GP has been widely applied to feature learning tasks, demonstrating strong potential particularly in classification problems \cite{tevc0,bitevc}. By embedding domain-specific knowledge into the evolutionary process, GP can automatically evolve explicit compositions of executable operations, producing features that are both effective and potentially interpretable.

Several representative works illustrate how explicitly incorporating such domain knowledge drives the success of GP across diverse classification tasks. In image classification, Bi et al. \cite{bitevc} integrate image-specific operators such as image filtering and pooling into a flexible GP representation, enabling the evolution of variable-depth trees that learn varying numbers and types of features. Similarly, Wang et al. \cite{wangtevc} develop a GP method for fine-grained image classification that explicitly incorporates visual prior knowledge into the search space. By guiding the evolutionary search to focus on object-aware local regions alongside color, shape, and texture cues, this approach improves both the discriminative power and interpretability of the learned features. In multimodal medical image classification, Wu et al. \cite{wutevc} propose a multitree GP framework that incorporates a slice selection factor based on the clinical insight that only a few image slices are diagnostically informative. The evolved individuals can be inspected to reveal which slices and anatomical regions are used and how modality-specific features are fused, enabling verification against established clinical biomarkers.

These studies demonstrate that, when equipped with suitable domain-specific knowledge, GP can automatically evolve effective and potentially interpretable feature learning models. While this highlights GP as a highly promising paradigm, research on GP-based feature learning specifically tailored to TSC remains very limited. The unique characteristics of time series data demand specialized program structures and tailored evolutionary search strategies. Accordingly, we develop a new GP method specifically for feature learning in TSC.

\section{Proposed Approach}
\label{s3}
In this study, we focus on feature learning for univariate time series classification. This work considers the offline time series classification setting, where each instance is a complete, equal-length sequence that has already been acquired and is classified as a whole. It does not address streaming classification, where data arrive incrementally and a decision may be required before the full sequence is observed. Let $D = \{(\bm{x}_i, y_i)\}_{i=1}^{N}$ denote a time series dataset, where $N$ is the number of instances, $\bm{x}_i \in \mathbb{R}^{L}$ is a series of length $L$, and $y_i$ is the corresponding class label.
The proposed EvoTSC aims to automatically evolve a feature learning model $\bm{\phi}(\cdot)$ that maps raw inputs into a discriminative feature space, i.e., $\bm{z}_i=\bm{\phi}(\bm{x}_i)\in\mathbb{R}^{d}$. 
The resulting features aim to highlight task-relevant information, maximizing class separability for a simple base classifier $h(\cdot)$ to achieve high generalization performance on unseen data.

In the following, Sections \ref{s31} to \ref{s33} describe how a single feature learning model is evolved by EvoTSC, and Section \ref{s3d} further describes how multiple independently evolved feature learning models and their base classifiers are combined into the proposed ensemble classifier EvoTSC-E.

\subsection{Algorithm Overview}
\label{s31}
\begin{figure}[!t]
 	\centering
 	\includegraphics[width=0.35\textwidth]{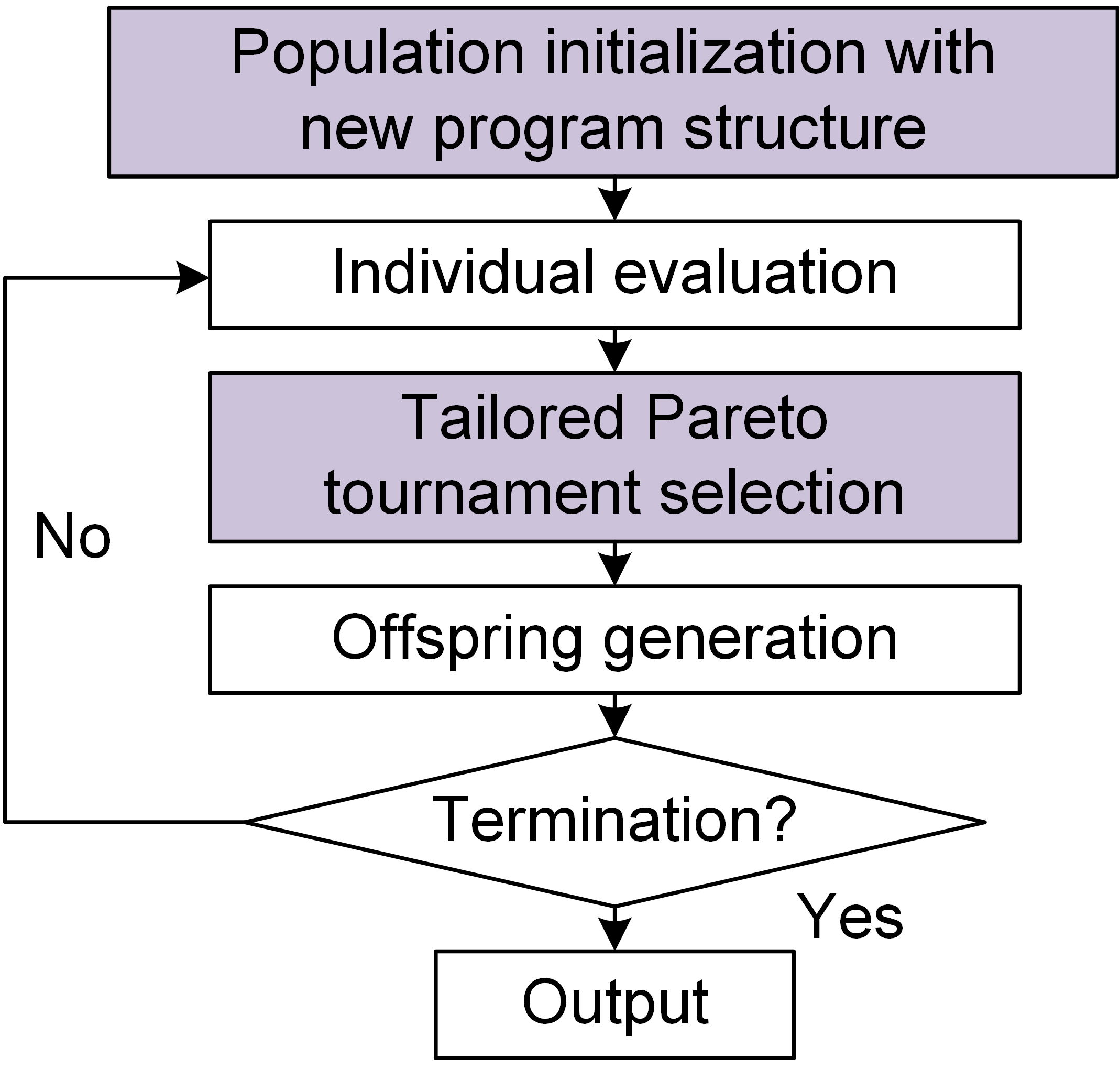}
 	\caption{Overall evolutionary process of the proposed GP method.}
 	\label{fig:as1}
 	\vspace{-2mm}
 \end{figure}

The overall evolutionary process of the proposed GP method is illustrated in Fig. \ref{fig:as1}, with the components newly designed for TSC highlighted in purple. It operates through an iterative loop comprising four main steps until a predefined maximum number of generations $G$ is reached or an individual attains 100\% cross-validation accuracy on the training set. These four stages are described below.
 
\subsubsection{\textbf{Population Initialization}}
At this stage, $m$ strongly typed GP trees \cite{stgp} are randomly generated using the ramped half-and-half strategy to initialize a population of $m$ individuals. This procedure combines “full” and “grow” constructions, thereby promoting both structural diversity and functional variety at the start of evolution. Primitives at different layers are sampled from layer-specific terminal or function sets, and every primitive is associated with an explicit input/output type signature. This enforces a predefined ordering of feature learning operations and guarantees that the resulting models are logically valid.

\subsubsection{\textbf{Individual Evaluation}}
Individual evaluation is performed on the training set using $k$-fold cross-validation to encourage generalization of the learned features, given that TSC tasks typically have limited labeled data.  Specifically, the training set is transformed according to the evolved individual (i.e., a feature learning model).  The fitness is then defined as the $k$-fold cross-validation result of an extremely randomized trees classifier \cite{geur006extremely} on the transformed training set. Following common practice in the GP-based feature learning community \cite{gpfan}, $k$ is set to $5$.

\subsubsection{\textbf{Parent Selection}}
The parent selection stage identifies high-quality, diverse individuals to ensure that effective building blocks transfer to the next generation. To prevent premature convergence and mitigate the common overfitting problem in TSC tasks, a tailored Pareto tournament selection strategy is employed. This approach is expected to enhance the generalization of the evolved feature learning models while preserving the localized competition characteristic of standard tournament selection, thereby maintaining population diversity. Further details of this strategy are provided in Section \ref{s33}.

\subsubsection{\textbf{Offspring Generation}}
The next generation of individuals is produced by applying elitism, crossover, and mutation with predefined probabilities. 
Elitism carries a small number of top-ranked individuals from the current population into the next generation unchanged, avoiding the loss of high-quality solutions. 
Using the selected parents, crossover and mutation are applied to generate the remaining individuals for the next generation. 
Together, these operations balance exploitation of high-quality structures and exploration of new program compositions. 

\subsection{Proposed Multi-layer Program Structure}
\label{s32}
\begin{figure*}[!t]
	\centering
	\includegraphics[width=1\textwidth]{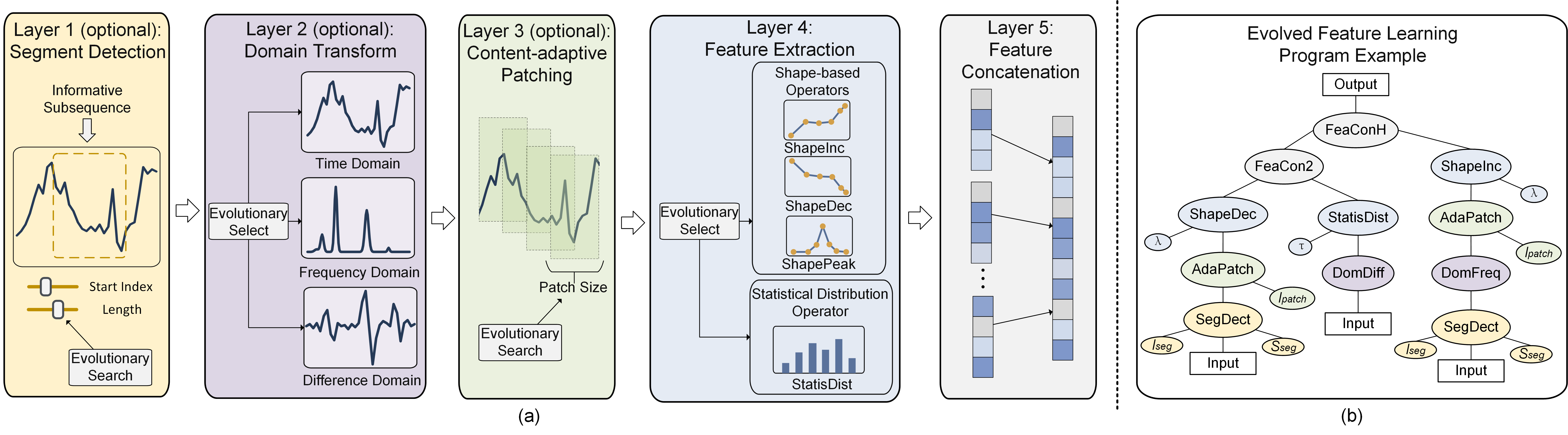}
	\vspace{-20pt}
	\caption{Illustration of the proposed multi-layer program structure (a) and an evolved feature learning model example (b).}
	\label{fig:evotsc_program_structure}
\end{figure*}

Although many GP methods have been proposed for feature learning in other problems, their program structures are not suited to TSC due to the unique characteristics of time series data. Therefore, this section proposes a new multi-layer program structure specifically designed for TSC. The proposed structure and an evolved feature learning model example are illustrated in Fig. \ref{fig:evotsc_program_structure}, and the details of each layer are described below.

\subsubsection{\textbf{Segment Detection Layer}}
\label{s321}
The first layer of the proposed structure is an optional segment detection layer, designed to identify an informative segment within a given time series. The segment detection operation is defined as:
\begin{equation}
	\begin{gathered}
		\operatorname{\mathbf{SegDect}}(\bm{x}) = (x_{s_{\text{seg}}}, x_{s_{\text{seg}}+1}, \ldots, x_{s_{\text{\text{seg}}}+\ell_{\text{seg}}-1}),
	\end{gathered}
\end{equation}
where $\ell_{\text{seg}}$ and $s_{\text{seg}}$ are two terminals for this layer, representing the segment length and the start index, respectively. The evolutionary search optimizes their values subject to $1\le \ell_{\text{seg}} < L$ and $1\le s_{\text{seg}}\le L-\ell_{\text{seg}}+1$. 

This layer is designed to provide practitioners with a clear and intuitive entry point for understanding the decision-making process. For example, in clinical electrocardiogram (ECG) analysis, rather than presenting the entire series, the evolved EvoTSC model isolates a specific segment, allowing domain experts to visually verify whether it corresponds to a meaningful physiological event, such as a transient arrhythmia or a specific QRS complex. Similarly, in motion classification, isolating the exact moment of a brief gesture makes the decision logic of the evolved EvoTSC model easy to inspect.

\subsubsection{\textbf{Domain Transform Layer}}
\label{s322}
Unlike many other data modalities, time series can often reveal salient class-discriminative information more explicitly after an appropriate domain transform. 
In the time domain, class differences driven by frequency characteristics may appear only as irregular oscillations. 
In contrast, frequency domain features often reveal these distinctions directly as shifts in the distribution of spectral energy. 
Furthermore, many classification tasks depend primarily on temporal changes rather than absolute values. 
Mapping a series into the difference-domain can highlight discriminative dynamics, such as rapid increases and slow decays, thereby supporting more effective classification. 

Motivated by these observations, the second layer is formulated as a domain transform layer that determines, in a data-driven manner, which domain yields the most informative features for classification. The function set of this layer involves two primitives: a frequency domain transform operator $\operatorname{\mathbf{DomFreq}}(\cdot)$ and a first-order differencing operator $\operatorname{\mathbf{DomDiff}}(\cdot)$. Let $\bm{x}_{\text{in}}=(x_1,x_2,\ldots,x_{\ell_{\text{in}}})\in\mathbb{R}^{\ell_{\text{in}}}$ denote the input segment passed from the preceding layer. The function $\operatorname{\mathbf{DomFreq}}(\cdot)$ computes the magnitude spectrum of $\bm{x}_{\text{in}}$ via the discrete Fourier transform (DFT):
\begin{equation}
    \begin{gathered}
	x'_{u}=\left|\sum_{t=0}^{\ell_{\text{in}}-1} x_{t+1}\exp\!\left(-i\,\frac{2\pi}{\ell_{\text{in}}}(u-1)t\right)\right|, \\
		\text{s.t.} \quad u=1,\ldots,\lfloor \ell_{\text{in}}/2\rfloor+1,
    \end{gathered}
\end{equation}
where $|\cdot|$ denotes the modulus. $\operatorname{\mathbf{DomDiff}}$ computes the first-order differences as:
\begin{equation}
\operatorname{\mathbf{DomDiff}}(\bm{x}_{\text{in}})=(x_2-x_1,\;x_3-x_2,\;\ldots,\;x_{\ell_{\text{in}}}-x_{\ell_{\text{in}}-1}).
\end{equation}
This layer is intentionally optional. When discriminative information is best captured in the time domain, the subsequence produced by the segment detection layer is passed directly to the next layer without applying a transform.

\subsubsection{\textbf{Content-adaptive Patching Layer}}
\label{s323}
Patching has recently emerged as an effective technique in time series analysis, which is initially explored with Transformer architectures \cite{patch}. More recently, several studies have shown that patching can also work well with other backbones, such as CNN and MLP \cite{patch1,patch2}, suggesting that redefining the basic extraction unit from the full series to a set of overlapping patches can enhance feature learning for time series. 

Motivated by these findings, patching is incorporated as an optional layer within the proposed program structure. Given an input segment $\bm{x}_{\text{in}}\in\mathbb{R}^{\ell_{\text{in}}}$, this operation segments $\bm{x}_{\text{in}}$ into $n_{\text{patch}}$ overlapping patches with size $\ell_{\text{patch}}$ and stride $s_{\text{patch}}$. The number of patches $n_{\text{patch}}$ is computed as:
\begin{equation}
	n_{\text{patch}} = \left\lfloor \frac{\ell_{\text{in}}-\ell_{\text{patch}}}{s_{\text{patch}}} \right\rfloor + 1.
\end{equation}
A practical challenge of patching is that the optimal patch size can vary substantially across datasets. 
Neural network approaches therefore often require per-dataset tuning to select an appropriate patch size. 
In contrast, the flexibility of GP provides a simple solution: by treating the patch size as a terminal, it can be optimized in a data-driven manner during evolution. 
The search space for $\ell_{\text{patch}}$ is restricted to $\{\lfloor \frac{\ell_{\text{in}}}{2}\rfloor,\lfloor \frac{\ell_{\text{in}}}{4}\rfloor,\ldots,\lfloor \frac{\ell_{\text{in}}}{64}\rfloor\}$. 
Moreover, following common practice in the prior study \cite{patch}, the stride is set to $s_{\text{patch}}=\lfloor \frac{\ell_{\text{patch}}}{2}\rfloor$.

\subsubsection{\textbf{Feature Extraction Layer}}
\label{s324}
After patching, the resulting set of $n_{patch}$ patches is passed to the feature extraction layer. This layer defines a set of extraction operators that produce discriminative patch-level features by explicitly capturing two complementary types of information crucial for TSC: shape-related characteristics and statistical distribution properties.

To extract shape information, convolution operations are particularly well-suited, as they measure the similarity between a local pattern and a template specified by a convolution kernel $\bm{w}$. For a processed patch $\bm{p}\in\mathbb{R}^{\ell_{\text{patch}}}$, the 1D convolution operation can be mathematically described as:
\begin{equation}
	\begin{gathered}
		r_j = (\bm{p} * \bm{w})_j = \sum_{i=0}^{c-1} p_{j+i} \cdot w_i, \\
		\text{s.t.} \quad j = 1, 2, \ldots, \ell_{\text{patch}} - c + 1,
	\end{gathered}
\end{equation}
where $\bm{r}$ is the activation map that highlights where and how strongly the pattern occurs within the processed segment, and $c$ is the kernel length.
A distinctive characteristic of time series data is that it can often be viewed as a composition of morphological primitives, such as increasing trends, decreasing trends, and peaks. Based on this observation, prior work \cite{craftkernel} proposes three predefined 1D convolution kernels designed to capture these specific patterns, thereby explicitly encoding shape information. Building upon these three predefined kernels\footnote{Due to space limitations, we omit the detailed formulations here. The explicit definitions of the three convolutional kernels can be found in \cite{craftkernel}.}, this layer defines a set of three corresponding feature extraction operators, denoted as $\mathcal{F}_{\text{shape}} = \{\operatorname{\mathbf{ShapeInc}}, \operatorname{\mathbf{ShapeDec}}, \operatorname{\mathbf{ShapePeak}}\}$, which capture increasing trends, decreasing trends, and peaks, respectively. During the evolutionary search, GP automatically determines which operator $\operatorname{\mathbf{F}}_{\text{shape}} \in \mathcal{F}_{\text{shape}}$ (shape prior) is most discriminative for the processed segment. Once an operator is selected, convolution for the chosen pattern is performed on the z-normalized patch using a set of kernel
lengths $\mathcal{C}=\{2^i \mid 2 \le 2^i < \lambda \cdot \ell_{\text{patch}}\}$, where $\lambda$ is a
terminal with the possible values $\{0.25, 0.5, 0.75\}$. 
This multi-scale design effectively captures both short-term fluctuations and longer-term trends. In \cite{craftkernel}, these predefined kernels are embedded within a CNN architecture, whereby the semantic clarity of the initial features is obscured by subsequent black-box transformations. To address this limitation, the selected operator $\operatorname{\mathbf{F}}_{\text{shape}}(\cdot)$ defined in this layer directly summarizes each resulting activation map $\bm{r}^{c}$ using three pooling operators, which capture complementary aspects of pattern occurrence:

\begin{align}
	\operatorname{PPV}(\bm{r}^{c}) &=
	\frac{1}{|\bm{r}^{c}|}\sum_{t=1}^{|\bm{r}^{c}|}\mathbb{I}\!\left[r^{c}_t>0\right],
	\label{eq:ppv_pool}
	\\
	\operatorname{MAX}(\bm{r}^{c}) &= \max\ \bm{r}^{c},
	\label{eq:max_pool}
	\\
	\operatorname{MEAN}(\bm{r}^{c}) &= \frac{1}{|\bm{r}^{c}|}\sum_{t=1}^{|\bm{r}^{c}|} r^{c}_t,
\end{align}
where $\mathbb{I}[\cdot]$ is the indicator function. $\operatorname{PPV}$ measures the frequency of activations, $\operatorname{MAX}$ captures the strongest match, and $\operatorname{MEAN}$ reflects the overall response intensity. Finally, the layer outputs a feature vector by concatenating pool descriptors across all kernel lengths:
\begin{equation}
	\operatorname{\mathbf{F}}_{\text{shape}}(\bm{p})
	=
	\bigoplus_{c\in\mathcal{C}}
	\Big(
	\operatorname{PPV}(\bm{r}^{c}),\;
	\operatorname{MAX}(\bm{r}^{c}),\;
	\operatorname{MEAN}(\bm{r}^{c})
	\Big),
\end{equation}
where $\oplus$ denotes vector concatenation. This yields features that are effective for classification while maintaining semantic clarity.

Beyond shape-related information, classes can also differ in the statistical distribution of the observed values. Classic summary statistics, such as the minimum, maximum, and median, provide simple and interpretable descriptors of a distribution. These statistics motivate a broader view: rather than extracting only a small set of fixed features, a discriminative feature can be constructed by sampling the empirical quantile function more densely, which captures the distribution profile of a segment \cite{ecdf, quant}. To capture this information, this layer defines an extraction operator for statistical distribution features, denoted as $\operatorname{\mathbf{StatisDist}}(\cdot)$. Given a processed patch $\bm{p}\in\mathbb{R}^{\ell_{\text{patch}}}$, its values are treated as an empirical sample from an unknown distribution. To form a distributional encoding, the values of $\bm{p}$ are sorted in ascending order to obtain the order statistics $\bm{p}^{\uparrow} = (p^{\uparrow}_1, p^{\uparrow}_2, \ldots, p^{\uparrow}_{\ell_{\text{patch}}})$. The sorted sequence is then uniformly subsampled to retain a proportion $\tau$ of the points ($M=\lceil \tau \times \ell_{\text{patch}} \rceil$), where $\tau$ is a terminal with the possible values $\{0.25, 0.5, 0.75\}$, allowing GP to automatically select the sampling density in a data-driven manner. Let the sampling positions be:
\begin{equation}
	idx_j = 1 + \frac{(\ell_{\text{patch}}-1)\,j}{M-1}, \quad j=0,\ldots,M-1,
\end{equation}
and this operator can be defined as:
\begin{equation}
	\operatorname{\mathbf{StatisDist}}(\bm{p}) = \big[p^{\uparrow}_{idx_0},\,p^{\uparrow}_{idx_1},\,\ldots,\,p^{\uparrow}_{idx_{M-1}}\big],
\end{equation}
where $p^{\uparrow}_{idx_j}$ for a non-integer $idx_j$ is computed by linear interpolation between the two adjacent order statistics. Meanwhile, the patch mean $\bar{p}$ is subtracted from every second quantile. This allows the features to capture both the value distribution and deviations from the mean without increasing the number of features \cite{quant}.

Ultimately, evolutionary search selects a suitable extraction operator $\mathbf{F} \in \{\mathbf{ShapeInc}, \allowbreak \mathbf{ShapeDec}, \allowbreak \mathbf{ShapePeak}, \allowbreak \mathbf{StatisDist}\}$. This selected operator is then applied identically and independently to all $n_{patch}$ patches. This shared application ensures that localized discriminative information can be captured regardless of where it occurs in the segment. Finally, the layer outputs the feature vector by concatenating the extraction results of all patches:
\begin{equation}
	\bm{z} = \bigoplus_{i=1}^{n_{patch}} \operatorname{\mathbf{F}}(\bm{p}_i).
\end{equation}

\subsubsection{\textbf{Feature Concatenation Layer}}
\label{s325}
While the preceding layers learn discriminative  features from specific segments or domains, a single learned feature vector branch may be insufficient for complex TSC tasks. 
Therefore, this final layer concatenates multiple feature vectors.

To support flexible multi-branch composition, this layer provides two categories of functions. 
The first category includes the base functions $\operatorname{\mathbf{FeaCon2}}$, $\operatorname{\mathbf{FeaCon3}}$, and $\operatorname{\mathbf{FeaCon4}}$, which concatenate two, three, or four feature vectors into a single output, respectively. 
The second category, $\operatorname{\mathbf{FeaConH}}$, concatenates a feature vector with the output of another concatenation function.

To provide a clear overview of the program structure, the function and terminal sets of the proposed multi-layer program structure are summarized in Table \ref{tab:function_set} and Table \ref{tab:terminal_set}. The proposed multi-layer structure  explores the inherent advantages of GP by embedding diverse forms of prior expert knowledge into the feature learning process. By doing so, the search space is purposefully biased toward operations known to be effective for TSC tasks. Ultimately, the evolutionary process aims to select and compose appropriate expert components from a unified search space according to the characteristics of the input data.

\begin{table}[htbp]
	\centering
	\caption{Function Set of the Proposed Multi-layer Program Structure.}
	\label{tab:function_set}
	\resizebox{\columnwidth}{!}{
		\begin{tabular}{@{}llll@{}}
			\toprule
			\textbf{Layer} & \textbf{Function} & \textbf{Input Type} & \textbf{Output Type} \\
			\midrule
			Segment Detection & $\operatorname{\mathbf{SegDect}}$ & Series, $\ell_{\text{seg}}$, $s_{\text{seg}}$ & Segment \\
			\midrule
			\multirow{2}{*}{Domain Transform} & $\operatorname{\mathbf{DomFreq}}$ & Series/Segment & Segment \\
			& $\operatorname{\mathbf{DomDiff}}$ & Series/Segment & Segment \\
			\midrule
			\makecell[l]{Content-adaptive\\Patching} & $\operatorname{\mathbf{AdaPatch}}$ & Series/Segment, $\ell_{\text{patch}}$ & Patches \\
			\midrule
			\multirow{4}{*}{\makecell[l]{Feature\\Extraction}} & $\operatorname{\mathbf{ShapeInc}}$ & Series/Segment/Patches, $\lambda$ & Vector \\
			& $\operatorname{\mathbf{ShapeDec}}$ &  Series/Segment/Patches, $\lambda$ & Vector \\
			& $\operatorname{\mathbf{ShapePeak}}$ &  Series/Segment/Patches, $\lambda$ & Vector \\
			& $\operatorname{\mathbf{StatisDist}}$ &  Series/Segment/Patches, $\tau$ & Vector \\
			\midrule
			\multirow{4}{*}{\makecell[l]{Feature\\Concatenation}} & $\operatorname{\mathbf{FeaCon2}}$ & 2 Vector & Vectors \\
			& $\operatorname{\mathbf{FeaCon3}}$ & 3 Vector & Vectors \\
			& $\operatorname{\mathbf{FeaCon4}}$ & 4 Vector & Vectors \\
			& $\operatorname{\mathbf{FeaConH}}$ & Vector, Vectors & Vectors \\
			\bottomrule
		\end{tabular}
	}
\end{table}

\begin{table}[htbp]
	\centering
	\caption{Terminal Set of the Proposed Multi-layer Program Structure.}
	\label{tab:terminal_set}
	\resizebox{\columnwidth}{!}{
		\begin{tabular}{@{}lll@{}}
			\toprule
			\textbf{Layer} & \textbf{Terminal} & \textbf{Search Range} \\
			\midrule
			\multirow{2}{*}{Segment Detection} & $\ell_{\text{seg}}$ & $\{1, 2, \ldots, L-1\}$ \\
			& $s_{\text{seg}}$ & $\{1, 2, \ldots, L-\ell_{\text{seg}}+1\}$ \\
			\midrule
			\makecell[l]{Content-adaptive\\Patching} & $\ell_{\text{patch}}$ & $\left\{\left\lfloor \frac{\ell_{\text{in}}}{2}\right\rfloor, \left\lfloor \frac{\ell_{\text{in}}}{4}\right\rfloor, \ldots, \left\lfloor \frac{\ell_{\text{in}}}{64}\right\rfloor\right\}$ \\
			\midrule
			\multirow{2}{*}{Feature Extraction} & $\tau$ & $\{0.25, 0.5, 0.75\}$ \\
			& $\lambda$ & $\{0.25, 0.5, 0.75\}$ \\
			\bottomrule
		\end{tabular}
	}
\end{table}

\subsection{Parent Selection Strategy}
\label{s33}
\begin{algorithm}[t]
	\caption{Tailored Pareto Tournament Parent Selection}
	\label{alg:pareto_tournament}
	\small
	\begin{algorithmic}[1]
		\Require Population $\mathcal{P}$, tournament ratio $\mu$, number of folds $k$, required number of parents $N_p$
		\Ensure Parent pool $\mathcal{M}$
		
		\State $\mathcal{M} \gets [\,]$
        \State $n_t \gets \operatorname{round}\!\left(\mu|\mathcal{P}|\right)$
		
		\While{$|\mathcal{M}| < N_p$}
		\State Randomly sample a tournament set $\mathcal{T} \subseteq \mathcal{P}$ with $|\mathcal{T}| = n_t$
		\State $\mathcal{ND} \gets \emptyset$
		\ForAll{$\bm{\phi}_i \in \mathcal{T}$}
		\State dominated $\gets$ \textbf{false}
		\ForAll{$\bm{\phi}_j \in \mathcal{T}, \bm{\phi}_j \neq \bm{\phi}_i$}
		\If{$\bm{\phi}_j \succ \bm{\phi}_i$}
		\State dominated $\gets$ \textbf{true}
		\State \textbf{break}
		\EndIf
		\EndFor
		\If{not dominated}
		\State $\mathcal{ND} \gets \mathcal{ND} \cup \{\bm{\phi}_i\}$
		\EndIf
		\EndFor
		\State Append all individuals in $\mathcal{ND}$ to $\mathcal{M}$ in a random order
		\EndWhile
		
		\State \Return first $N_p$ individuals in $\mathcal{M}$
	\end{algorithmic}
\end{algorithm}

Tournament selection used in previous GP-based feature learning methods \cite{bitevc,wangtevc,tevc0} typically relies on  the fitness value, i.e., the averaged
accuracy of $k$-fold cross-validation,  as the selection criterion. However, this scalar aggregation obscures performance differences across individual folds and allows strong performance on only one or two folds to compensate for poor performance on others. Consequently, it may misdirect the evolutionary process toward models that overfit to specific data splits, resulting in features with poor generalization ability. To mitigate this issue, a tailored Pareto tournament selection strategy \cite{paretotournament} is proposed.

Unlike standard Pareto tournament selection, which typically uses performance and model size as dual objectives, the proposed strategy treats the accuracy on each fold as an independent optimization objective. 
Specifically, the selection criterion is defined as a multi-objective vector:
\begin{equation}
	\bm{V}(\bm{\phi}) = \big( \operatorname{acc}_1(\bm{\phi}), \operatorname{acc}_2(\bm{\phi}), \ldots, \operatorname{acc}_k(\bm{\phi}) \big),
\end{equation}
where $\operatorname{acc}_j(\bm{\phi})$ denotes the classification accuracy on the $j$-th fold of the cross-validation performed on the training set in fitness evaluation.
During each selection event, a tournament is formed by randomly sampling a proportion $\mu$ of the population. 
This design retains the localized competition property of standard tournament selection, which helps preserve population diversity. Within this sampled subset, individuals are compared using Pareto dominance. 
An individual $\bm{\phi}_A$ is said to dominate $\bm{\phi}_B$ if and only if it performs no worse than $\bm{\phi}_B$ across all $k$ folds and performs strictly better on at least one fold. Unlike standard tournament selection, which selects only a single winner per event, the proposed strategy leverages this dominance relation to select the local non-dominated set from the tournament. All individuals within this non-dominated set are then added to the parent pool, and the tournament process is repeated until the required number of parents is obtained. An individual may be added multiple times if it is non-dominated in multiple tournaments. The complete procedure of the tailored Pareto tournament selection is summarized in Algorithm \ref{alg:pareto_tournament}. 

By enforcing Pareto dominance across all data splits, the selection mechanism creates evolutionary pressure that favors feature learning models capable of learning universally stable features across varying data subsets.

\subsection{Ensemble Classifier: EvoTSC-E}
\label{s3d}
Ensemble learning has demonstrated strong effectiveness in TSC. For example, InceptionTime combines multiple independently trained Inception networks \cite{inceptiontime}, while HIVE-COTE  combines multiple heterogeneous time series classifiers \cite{hivecote}. Meanwhile, the feature learning models evolved by the proposed EvoTSC method can be easily combined into an ensemble. Therefore, this section investigates whether an ensemble classifier built upon the evolved feature learning models can further improve the classification performance, leading to the proposed EvoTSC-E method.

Specifically, the GP method described in Sections \ref{s31} to \ref{s33} is executed $n_E$ times independently, where all runs adopt identical parameter settings and differ only in the random seed. Each run outputs its best evolved feature learning model, yielding a set of models $\{\phi_1, \phi_2, \ldots, \phi_{n_E}\}$. Since different models produce feature vectors of different dimensionalities, a base classifier $h_j$ is independently trained for each $\phi_j$ on the correspondingly transformed training set $\{(\phi_j(\boldsymbol{x}_i), y_i)\}_{i=1}^{N}$. Given an unseen instance $\boldsymbol{x}$, each base classifier produces a predicted class probability distribution over the label set $\mathcal{Y}$. Following the ensemble scheme of InceptionTime \cite{inceptiontime}, the final prediction is obtained by averaging the probability distributions of all base classifiers:
\begin{equation}
	\hat{y} = \arg\max_{c \in \mathcal{Y}} \frac{1}{n_E} \sum_{j=1}^{n_E} \big[ h_j(\phi_j(\boldsymbol{x})) \big]_c ,
\end{equation}
where \([\cdot]_c\) denotes the predicted probability associated with class \(c\).

Since EvoTSC-E incurs additional computational cost in the inference stage, it is intended for application scenarios in which computational resources are less constrained and classification accuracy is the primary concern.

\section{Experiment Design}
\label{s4}
\subsection{Datasets}
\label{s41}
Rather than relying on additional resampling or imputation procedures that may introduce confounding factors, we focus on datasets whose time series are equal-length and free of missing values. Accordingly, the experiments are conducted on 86 time series datasets from the UCR Time Series Classification Archive \cite{UCRArchive2018}. A detailed summary of these datasets is provided in Section \Romannum{1} of the supplementary material. The total number of instances per dataset ranges from 40 to 5{,}000. Throughout the experiments, both the proposed methods and all baselines use the original train/test data splits provided by the archive, with no additional normalization applied beyond those already incorporated into the archived datasets.

These datasets exhibit considerable diversity in their intrinsic characteristics: the number of classes ranges from 2 to 60, and the time series length varies from 15 to 2{,}800. Moreover, they are collected from a broad spectrum of application scenarios, including biomedical (10), industrial (21), motion (17), image outline (20), spectral (11) and simulated (7), providing a comprehensive testbed for evaluating the effectiveness of the learned features.

\subsection{Comparison Methods}
\label{s42}
The comparison methods are organized into two groups according to the distinct objectives of the experiments. The first group is used to evaluate the effectiveness of the feature learning models evolved by EvoTSC, whereas the second group is used to evaluate the classification performance of the EvoTSC-E method.

\subsubsection{Methods for Evaluating the Evolved Feature Learning Models}
To evaluate whether the learned features improve the classification performance, five common machine learning classifiers are employed: $k$-nearest neighbour ($k$-NN), support vector machine (SVM), logistic regression (LR), random forest (RF), and extremely randomized trees (ET). They are selected because they are widely used in classification and cover linear, distance-based, and tree-based learning paradigms. Furthermore, several representative feature-based TSC methods are included: the hand-crafted feature extraction method catch22 \cite{catch22}; the NN-based feature learning methods TS2Vec \cite{ts2vec} and SoftCLT \cite{softclt}, which are trained on the target dataset; and the foundation model-based methods MOMENT \cite{moment}, Mantis \cite{mantis}, and TiViT \cite{tivit}, which are evaluated under a zero-shot protocol, without fine-tuning on the target datasets.

\subsubsection{Methods for Evaluating EvoTSC-E}
Since EvoTSC-E directly produces classification results by combining multiple evolved feature learning models and their base classifiers, it is compared with a broader range of TSC methods rather than only feature-based methods. The comparison methods cover four families: the distance-based method 1NN-DTW \cite{dtw}, the dictionary-based method WEASEL \cite{weasel}, the random convolutional kernel-based methods ROCKET \cite{rocket} and MiniRocket \cite{minirocket}, and the ensemble methods InceptionTime \cite{inceptiontime} and HIVE-COTE 1.0 (HC1) \cite{hivecote}. Since ROCKET and MiniRocket are non-ensemble methods, we additionally construct their ensemble variants, ROCKET-E and MiniRocket-E, for comparisons with EvoTSC-E. We also include a single Inception network as a non-ensemble counterpart to InceptionTime. For fair comparisons, EvoTSC-E, ROCKET-E, and MiniRocket-E adopt the same ensemble size and probability-averaging strategy as InceptionTime. Specifically, each ensemble consists of five independently constructed base models, and the final class probability distribution is obtained by averaging the probability distributions predicted by the five base classifiers. Together, these methods provide a strong and representative benchmark for assessing the classification performance of EvoTSC-E.

Due to the page limit, detailed descriptions, implementation sources, and hyperparameter settings of all comparison methods are provided in Section \Romannum{2} of the supplementary material. For deterministic  methods, only one run is performed, as repeated runs yield identical results. For stochastic comparison methods and EvoTSC, 30 runs are performed using 30 different random seeds. Each EvoTSC-E run consists of five independent EvoTSC runs. Therefore, 150 EvoTSC runs with 150 different random seeds are performed on each dataset to obtain 30 EvoTSC-E runs. The same procedure is applied to ROCKET-E and MiniRocket-E.

The statistical significance of performance differences is evaluated on each dataset at a significance level of $\alpha = 0.05$. The choice of statistical significance test depends on whether the methods are stochastic or deterministic. For comparisons between two stochastic methods, the 30 runs of experiments are paired based on the same random seeds, and the Wilcoxon signed-rank test is applied to the resulting 30 accuracy differences. For comparisons between a stochastic and a deterministic method, the $t$-test is used to compare the mean accuracy of the stochastic method with the fixed accuracy of the deterministic method. For comparisons between two deterministic methods, only their accuracies are compared, without a statistical significance test. To account for multiple comparisons, Holm's correction is applied to the 86 dataset-level $p$-values for each comparison, where applicable.

\subsection{Parameter Settings of the Proposed GP Method}
\label{s43}
Following the common practice in GP-based feature learning methods \cite{wutevc}, the parameter settings of the proposed EvoTSC method are shown in Table \ref{tab:gp_params}. The minimum tree depth is set to 2, corresponding to the minimal model that employs only the feature extraction and feature concatenation layers as fixed components. The maximum tree depth is set to 6, corresponding to the maximal model that activates all layers and utilizes $\operatorname{\mathbf{FeaConH}}$ for feature concatenation. Since fitness evaluation in GP mainly provides a relative ranking signal for comparing candidate feature learning models rather than a precise estimate of absolute performance differences \cite{9656554,10783045}, a lightweight classifier is sufficient for this purpose. Based on preliminary experiments, an extremely randomized trees classifier with 10 trees is adopted, as it achieves a good balance between evaluation efficiency and ranking reliability.

\begin{table}[t]
	\centering
	\caption{Parameter Settings of the Proposed Method.}
	\label{tab:gp_params}
	\renewcommand{\arraystretch}{1.15}
	\begin{tabular}{c c}
		\hline
		\textbf{Parameter} & \textbf{Value} \\
		\hline
		Population size & 100 \\
		Maximum number of generations & 50 \\
		Crossover and mutation rates & 0.8 and 0.19 \\
		Elitism rate & 0.01 \\
		Minimum tree depth & 2 \\
		Maximum tree depth & 6 \\
		\hline
	\end{tabular}
\end{table}

\section{Results and Analysis}
\label{s5}
\subsection{Effectiveness of the Learned Features}
\label{s51}
In this section, the effectiveness of the learned features is investigated by comparing the performance of common machine learning classifiers applied directly to raw time series data against their performance when coupled with the evolved feature learning models. The results of the statistical significance tests are summarized as W/T/L counts, representing the number of datasets on which EvoTSC performs significantly better than, shows no significant difference from, and performs significantly worse than the raw data baseline.

\begin{figure}
	\centering
	\includegraphics[width=0.8\columnwidth]{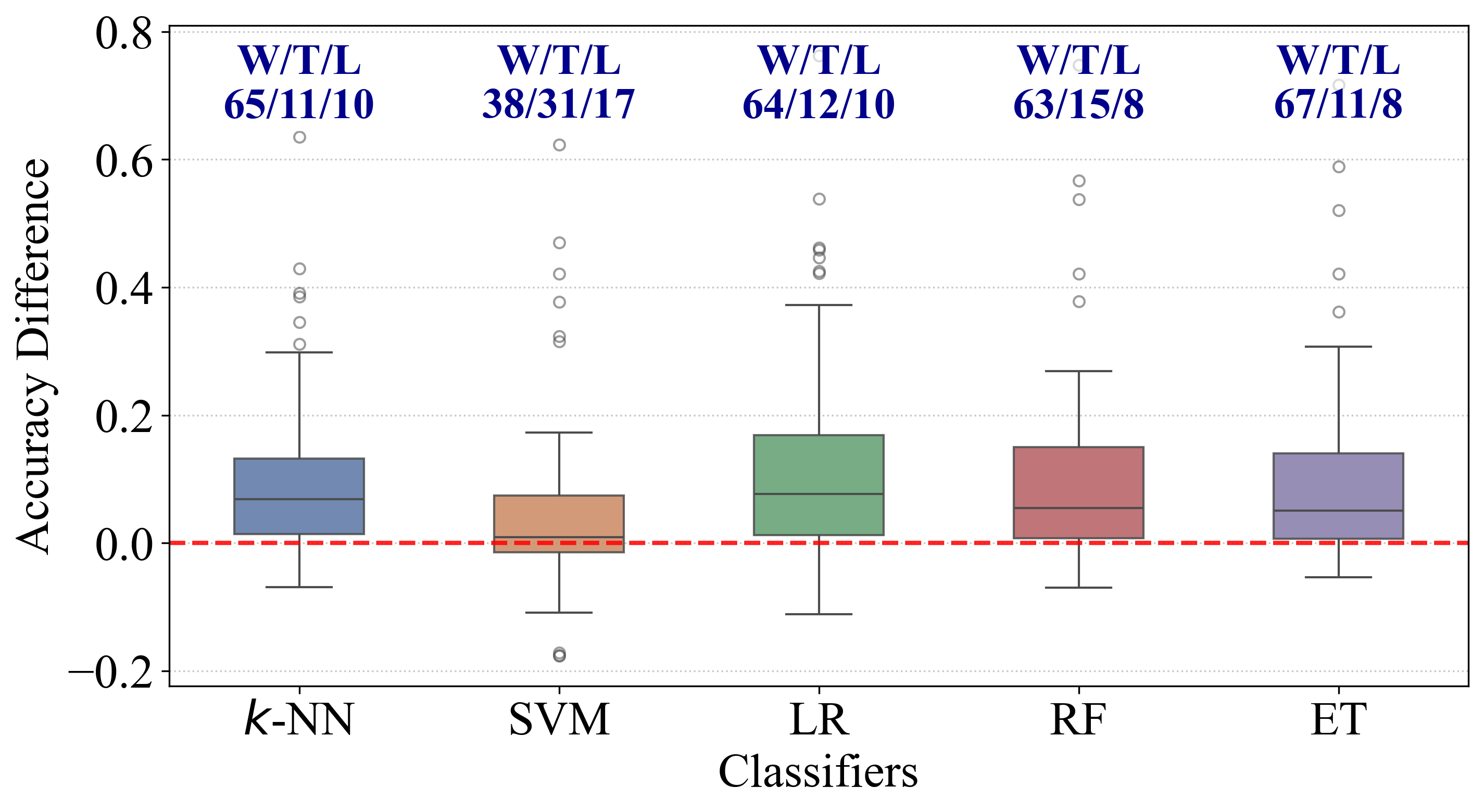}
	\vspace{-10pt}
	\caption{Distribution of accuracy improvement over the raw data baseline across all datasets. The W/T/L counts annotated above each box denote the numbers of datasets on which EvoTSC performs significantly better than (W), shows no significant difference from (T), and performs significantly worse than (L) the raw data baseline, according to the statistical significance tests. The red dashed line at $y=0$ denotes no accuracy improvement over the raw data baseline.}
	\label{fig:effectiveness}
\end{figure}

As shown in Fig. \ref{fig:effectiveness}, the W/T/L counts indicate that the features learned by EvoTSC significantly outperform the raw data or show no significant difference from it on over 80\% of the datasets for every classifier. A number of positive outliers can be observed, indicating that on certain datasets the learned features bring particularly notable performance improvement. While a small number of negative outliers also exist, they are considerably fewer.

\subsection{Comparisons with Feature-based TSC Methods}
\label{s52}
This section evaluates the evolved feature learning models through comparisons with six feature-based TSC methods. The models evolved by EvoTSC use the ET classifier, as it yields the largest number of datasets with significant improvement. For TS2Vec, SoftCLT, MOMENT, Mantis, and TiViT, the classifiers recommended in their respective original papers are employed. Since catch22 does not specify a recommended classifier, the same ET classifier as EvoTSC is used.

The experimental results are presented in Table \ref{tab:pairwise_sig}, where each cell reports the number of datasets on which the row method performs significantly better than, shows no significant difference from, and performs significantly worse than the column method.
Three observations can be made. First, EvoTSC shows a clear advantage over the hand-crafted feature extraction method catch22, with 68 wins and only six losses, demonstrating the benefit of adapting the feature learning process to each dataset. Second, compared with the self-supervised NN-based methods, EvoTSC achieves 30 and 39 wins against SoftCLT and TS2Vec, with 21 and 25 losses, respectively. Third, for the three TSFMs evaluated under the zero-shot protocol, EvoTSC significantly outperforms each of them on at least 41 of the 86 datasets. These results indicate that the dataset-specific feature learning models evolved by EvoTSC provide more discriminative features than those produced by the compared TSFMs on more datasets, despite the latter benefiting from large-scale external pre-training.
\begin{table}[ht]
	\centering
		\caption{Statistical comparison results across all datasets. Each cell reports the number of datasets on which the row method performs \textbf{significantly better} than, shows \textbf{no significant difference} from, and performs \textbf{significantly worse} than the column method.}
	\label{tab:pairwise_sig}
	\renewcommand{\arraystretch}{1.15}
	\setlength{\tabcolsep}{2pt}
	\scriptsize
	\begin{tabular}{lcccccc}
		\toprule
		\textbf{Method} & \textbf{TiViT} & \textbf{Mantis} & \textbf{MOMENT} & \textbf{SoftCLT} & \textbf{TS2Vec} & \textbf{catch22} \\
		\midrule
		EvoTSC  & 41/19/26 & 53/22/11 & 51/24/11 & 30/35/21 & 39/22/25 & 68/12/6  \\
		TiViT   & --       & 59/5/22  & 62/4/20  & 36/12/38 & 37/12/37 & 59/6/21  \\
		Mantis  & --       & --       & 42/7/37  & 19/18/49 & 20/13/53 & 47/10/29 \\
		MOMENT  & --       & --       & --       & 18/13/55 & 21/18/47 & 47/11/28 \\
		SoftCLT & --       & --       & --       & --       & 20/60/6  & 64/7/15  \\
		TS2Vec  & --       & --       & --       & --       & --       & 59/13/14 \\
		\bottomrule
	\end{tabular}
\end{table}

Furthermore, to rule out the possibility that the superior performance of EvoTSC is attributable to the choice of the ET classifier rather than the effectiveness of the evolved feature learning models, an additional experiment is conducted in which all comparison methods (excluding catch22, which already uses ET) use the same ET classifier. The results presented in Table \ref{tab:pairwise_et} show that EvoTSC continues to outperform all comparison methods under this unified classifier setting.
\begin{table}[ht]
	\centering
	\caption{Statistical comparison results under a unified ET classifier across all datasets. Each cell reports the number of datasets on which the row method performs \textbf{significantly better} than, shows \textbf{no significant difference} from, and performs \textbf{significantly worse} than the column method.}
	\label{tab:pairwise_et}
	\renewcommand{\arraystretch}{1.15}
	\setlength{\tabcolsep}{2pt}
	\scriptsize
	\begin{tabular}{lccccc}
		\toprule
		\textbf{Method} & \textbf{TiViT\_ET} & \textbf{Mantis\_ET} & \textbf{MOMENT\_ET} & \textbf{SoftCLT\_ET} & \textbf{TS2Vec\_ET} \\
		\midrule
		EvoTSC      & 49/22/15 & 52/18/16 & 62/17/7  & 46/28/12 & 45/32/9  \\
		TiViT\_ET   & --       & 42/13/31 & 62/9/15  & 34/15/37 & 35/13/38 \\
		Mantis\_ET  & --       & --       & 50/8/28  & 26/21/39 & 27/23/36 \\
		MOMENT\_ET  & --       & --       & --       & 18/15/53 & 17/19/50 \\
		SoftCLT\_ET & --       & --       & --       & --       & 19/59/8  \\
		\bottomrule
	\end{tabular}
\end{table}

\subsection{Grouped Dataset Analysis}
\label{s53}
To further investigate the performance of the evolved feature learning models across different dataset characteristics, this section conducts a group-wise analysis of the seven feature-based methods compared in Section \ref{s52}. The datasets are grouped according to three properties: the number of classes (four groups), time series length (four groups), and dataset type (six groups)\footnote{Detailed information on these properties for each dataset is provided in Section \Romannum{1} of the supplementary material.}. For each of the 14 resulting groups, the average rank of each method is computed across the corresponding datasets, as reported in Table \ref{tab:grouped}.

\subsubsection{Effect of the Number of Classes}
The upper panel of Table \ref{tab:grouped} partitions the datasets by the number of classes. When the number of classes is small ($C\!=\!2$), SoftCLT achieves the best average rank of 3.283, with EvoTSC following closely in second place (3.350). In the middle range, EvoTSC demonstrates a clear advantage: it achieves the best rank in both the $3 \!\leq\! C \!\leq\! 5$ group (2.852) and the $6 \!\leq\! C \!\leq\! 10$ group (2.333). In the many-class group ($C\!>\!10$), the NN-based feature learning methods TS2Vec (2.643) and SoftCLT (2.714) attain the two best ranks, with EvoTSC following closely in third place (3.000). Overall, EvoTSC is the only method that stays within the top three across all four class-count groups.

\subsubsection{Effect of Time Series Length}
The middle panel of Table \ref{tab:grouped} groups the datasets by the series length. EvoTSC ranks within the top two in three of the four length groups. Its rank of 2.471 on series exceeding length 1000 is its best among the four length groups. This can be attributed to the inherent flexibility of GP, which enables the production of a sufficiently rich feature set to capture the complex discriminative patterns embedded in long series.

\subsubsection{Effect of the Dataset Type}
The lower panel of Table \ref{tab:grouped} groups the datasets by application type. EvoTSC achieves the best average rank in three of the six types (Biomedical, ImageOutline, and Spectral). For Motion datasets, TiViT achieves the best rank (2.471), which is consistent with its design: by converting time series into images and exploiting a vision Transformer, it is naturally well suited to data characterized by action or trajectory-related local pattern changes. Similarly, TiViT ranks first on Simulated datasets (2.143), because many UCR simulated problems are built around embedded shape patterns. By contrast, TiViT ranks only fifth (4.275) on ImageOutline datasets. A possible explanation is that these datasets are already one-dimensional encodings of object outlines, so transforming them again into images may provide less additional benefit than methods that operate directly on the original 1D shape signal.

Overall, EvoTSC ranks within the top two in ten of the fourteen groups, more than any other compared method. These results indicate that EvoTSC can evolve effective feature learning models across datasets with diverse characteristics, highlighting its adaptability to different TSC scenarios.
\begin{table}[t]
	\centering
	\scriptsize
	\setlength{\tabcolsep}{2.2pt}
	\renewcommand{\arraystretch}{1.10}
	\caption{Average ranks of the seven feature-based methods within fourteen dataset groups. The best rank in each row is in \textbf{bold}, and the second best is \underline{underlined}. $n$ denotes the number of datasets in each group.}
	\label{tab:grouped}
	\resizebox{\columnwidth}{!}{
		\begin{tabular}{llccccccc}
			\toprule
			\multirow{2}{*}{Criterion} & \multirow{2}{*}{Group} &
			\multicolumn{7}{c}{Average Rank} \\
			\cmidrule(lr){3-9}
			& & catch22 & TS2Vec & SoftCLT & MOMENT & Mantis & TiViT & EvoTSC \\
			\midrule

			\multirow{4}{*}{Classes $C$}
			& $C=2$ ($n=30$)
			& 5.367 & 3.500 & \textbf{3.283} & 4.517 & 4.633 & \underline{3.350} & \underline{3.350} \\

			& $3\leq C\leq5$ ($n=27$)
			& 5.389 & 4.000 & \underline{3.444} & 4.241 & 4.389 & 3.685 & \textbf{2.852} \\

			& $6\leq C\leq10$ ($n=15$)
			& 5.200 & \underline{3.400} & 3.600 & 4.600 & 5.400 & 3.467 & \textbf{2.333} \\

			& $C>10$ ($n=14$)
			& 5.643 & \textbf{2.643} & \underline{2.714} & 6.000 & 4.714 & 3.286 & 3.000 \\
			\midrule

			\multirow{4}{*}{Length $L$}
			& $L\leq150$ ($n=31$)
			& 5.726 & 3.210 & \underline{2.871} & 4.645 & 4.758 & 4.081 & \textbf{2.710} \\

			& $151\leq L\leq500$ ($n=22$)
			& 6.136 & 3.568 & \underline{3.205} & 3.864 & 4.659 & \textbf{3.136} & 3.432 \\

			& $501\leq L\leq1000$ ($n=16$)
			& 4.813 & 4.063 & 3.875 & 5.125 & 4.031 & \textbf{2.781} & \underline{3.313} \\

			& $L>1000$ ($n=17$)
			& 4.353 & \underline{3.412} & 3.647 & 5.412 & 5.294 & \underline{3.412} & \textbf{2.471} \\
			\midrule

			\multirow{6}{*}{Type}
			& Biomedical ($n=10$)
			& 4.450 & 3.000 & \underline{2.800} & 6.150 & 4.500 & 4.400 & \textbf{2.700} \\

			& Industrial ($n=21$)
			& 4.500 & 4.048 & 3.714 & 5.048 & 5.000 & \textbf{2.833} & \underline{2.857} \\

			& Motion ($n=17$)
			& 5.794 & \underline{3.382} & 3.824 & 4.618 & 4.441 & \textbf{2.471} & 3.471 \\

			& ImgOutline ($n=20$)
			& 6.200 & 3.500 & \underline{3.000} & 3.975 & 4.300 & 4.275 & \textbf{2.750} \\

			& Spectral ($n=11$)
			& 5.545 & 3.409 & \underline{2.955} & 3.409 & 5.273 & 4.727 & \textbf{2.682} \\

			& Simulated ($n=7$)
			& 5.857 & 3.000 & \underline{2.857} & 5.714 & 5.000 & \textbf{2.143} & 3.429 \\
			\bottomrule
		\end{tabular}}
\end{table}

\subsection{Comparisons with Other TSC Methods}
\label{s5D}
\begin{figure}[ht]
	\centering
	\includegraphics[width=0.5\textwidth]{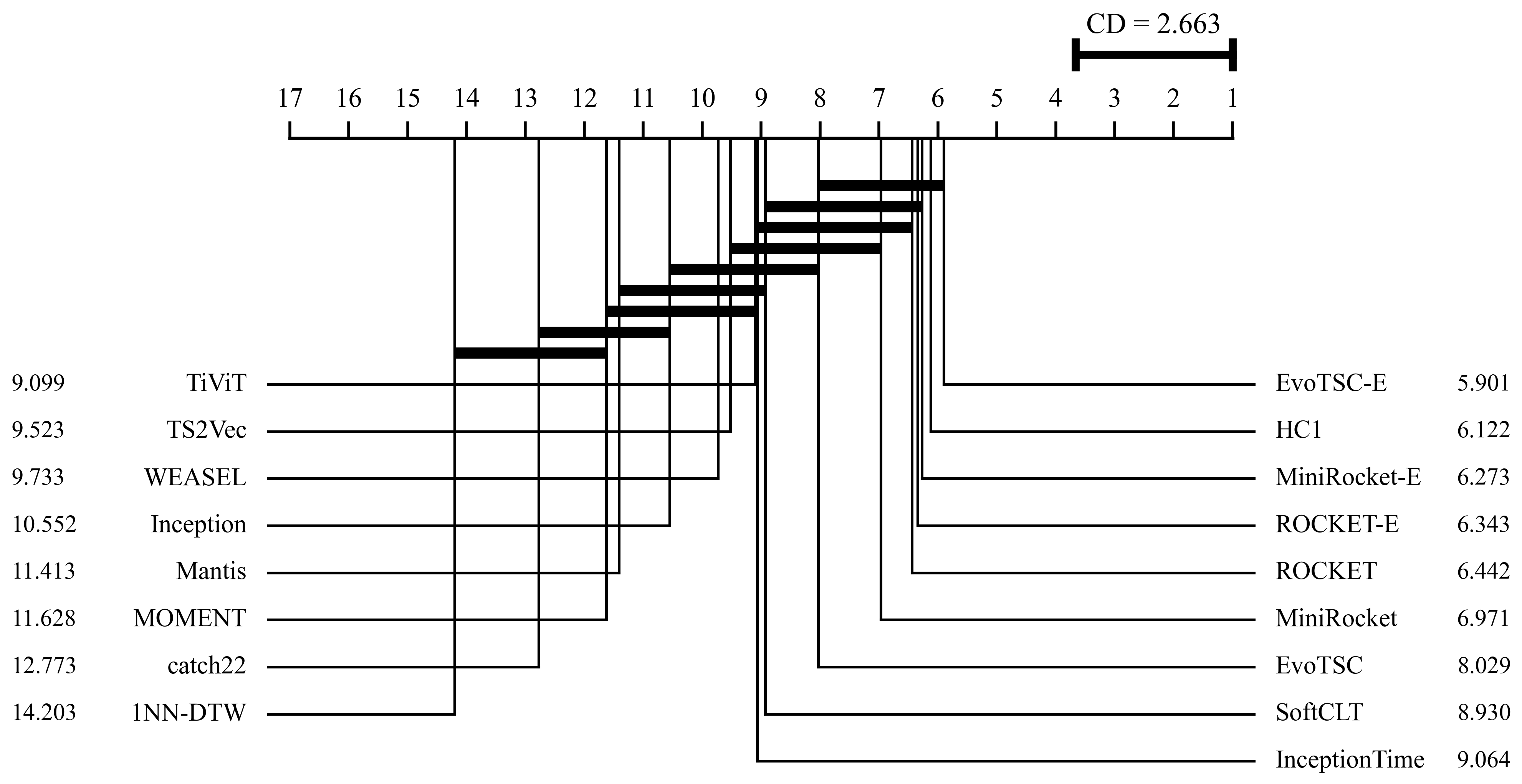}
	\vspace{-20pt}
	\caption{Critical difference diagram comparing all the 17 methods over 86 datasets. The CD is calculated by the Nemenyi post-hoc test at $\alpha=0.05$ and represents the minimum difference between two average ranks that can be regarded as statistically significant. Methods separated by less than the CD are connected by a thick horizontal line and form a clique.}
	\label{fig_cd_e}
\end{figure}

This section evaluates the classification performance of the proposed ensemble classifier EvoTSC-E against representative TSC methods. Table \ref{tab:pairwise_sota} reports the statistical comparison results between EvoTSC-E and the nine comparison methods over the 86 datasets. EvoTSC-E achieves more wins than losses against eight methods and equal wins and losses against HC1. Moreover, the critical difference (CD) diagram in Fig. \ref{fig_cd_e} compares EvoTSC-E, the non-ensemble EvoTSC, and all other compared methods using the Friedman test followed by the Nemenyi post-hoc test at a significance level of $\alpha=0.05$. It can be seen that EvoTSC-E achieves the best average rank of 5.901 among all the 17 methods. These results show that when computational resources are less constrained, multiple evolved feature learning models can be combined into a strong ensemble classifier.

For comparison, the non-ensemble EvoTSC achieves an average rank of 8.029, better than those of all compared feature-based methods and the ensemble-based InceptionTime. Although EvoTSC obtains a slightly worse rank than ROCKET and MiniRocket, no statistically significant difference can be detected among them, as all three methods fall within the same Nemenyi clique. Notably, EvoTSC achieves this level of performance with substantially lower computational cost. As shown in Section \ref{s63}, the model evolved by EvoTSC requires substantially fewer operations and less peak activation memory than ROCKET and MiniRocket. Moreover, ROCKET and MiniRocket extract features using large collections of randomly generated convolutional kernels. In contrast, each component of an evolved EvoTSC model has a predefined role, which gives the models greater inspectability. The per-dataset classification accuracies of all the methods compared in Fig. \ref{fig_cd_e} are provided in Section \Romannum{6} of the supplementary material.
\begin{table*}[ht]
	\centering
	\caption{Statistical comparison results across all datasets. Each cell reports the number of datasets on which the row method performs \textbf{significantly better} than, shows \textbf{no significant difference} from, and performs \textbf{significantly worse} than the column method.}
	\label{tab:pairwise_sota}
	\renewcommand{\arraystretch}{1.15}
	\setlength{\tabcolsep}{2pt}
	\scriptsize
	\begin{tabular}{lccccccccc}
		\toprule
		\textbf{Method} & \textbf{MiniRocket-E} & \textbf{ROCKET-E} & \textbf{HC1} & \textbf{InceptionTime} & \textbf{MiniRocket} & \textbf{ROCKET} & \textbf{Inception} & \textbf{WEASEL} & \textbf{1NN-DTW} \\
		\midrule
		EvoTSC-E      & 33/22/31 & 34/20/32 & 29/28/29 & 53/10/23 & 34/21/31 & 34/20/32 & 55/11/20 & 55/13/18 & 74/10/2 \\
		MiniRocket-E  & --       & 35/21/30 & 32/11/43 & 49/16/21 & 7/79/0   & 36/19/31 & 52/16/18 & 52/12/22 & 74/5/7  \\
		ROCKET-E      & --       & --       & 29/22/35 & 51/8/27  & 31/23/32 & 0/86/0   & 52/15/19 & 56/10/20 & 74/6/6  \\
		HC1           & --       & --       & --       & 48/12/26 & 43/12/31 & 35/24/27 & 52/13/21 & 62/11/13 & 75/7/4  \\
		InceptionTime & --       & --       & --       & --       & 20/20/46 & 27/8/51  & 16/70/0  & 41/8/37  & 60/8/18 \\
		MiniRocket    & --       & --       & --       & --       & --       & 33/22/31 & 50/18/18 & 51/13/22 & 73/5/8  \\
		ROCKET        & --       & --       & --       & --       & --       & --       & 52/17/17 & 55/12/19 & 74/6/6  \\
		Inception     & --       & --       & --       & --       & --       & --       & --       & 33/16/37 & 57/8/21 \\
		WEASEL        & --       & --       & --       & --       & --       & --       & --       & --       & 64/5/17 \\
		\bottomrule
	\end{tabular}
\end{table*}

\section{Further Analysis}
\label{s6}
\subsection{Ablation Studies}
\label{s61}
To investigate the contribution of each component in the proposed EvoTSC method, a comprehensive ablation study is conducted by systematically disabling individual components to produce degraded variants. Specifically, the ablation is organized from three perspectives, yielding nine degraded versions in total. Following the same grouped dataset analysis protocol as in Section \ref{s53}, the average rank of each variant is computed within each dataset group. The results are reported in Table \ref{tab:ablation}. 
\begin{table*}
	\centering
	\footnotesize
	\setlength{\tabcolsep}{3.0pt}
	\caption{Average rank of each ablation variant within fourteen dataset groups. The best rank in each row is in \textbf{bold} and the second best is \underline{underlined}. $n$ denotes the number of datasets in each group.}
	\label{tab:ablation}
	\resizebox{\textwidth}{!}{
		\begin{tabular}{llcccccccccc}
			\toprule
			\multirow{2}{*}{Criterion} & \multirow{2}{*}{Group} &
			\multicolumn{10}{c}{Average rank} \\
			\cmidrule(lr){3-12}
			& &
			ShapeInc & ShapeDec & ShapePeak & StatisDist & AllFea &
			wo\_Dom & wo\_Patch & wo\_Con & wo\_Pareto & \textsc{EvoTSC} \\
			\midrule
			\multirow{4}{*}{Classes $C$}
			& $C=2$ ($n=30$)
			& 7.933 & 8.067 & 6.567 & 7.717 & 5.183
			& 4.150 & 4.200 & 4.867 & \underline{3.633} & \textbf{2.683} \\
			& $3\leq C\leq5$ ($n=27$)
			& 8.370 & 8.130 & 6.704 & 8.019 & 6.222
			& 3.741 & 4.204 & 4.148 & \underline{2.759} & \textbf{2.704} \\
			& $6\leq C\leq10$ ($n=15$)
			& 7.400 & 8.067 & 8.200 & 8.467 & 5.733
			& 3.700 & 3.900 & 5.000 & \textbf{2.200} & \underline{2.333} \\
			& $C>10$ ($n=14$)
			& 8.929 & 8.786 & 7.571 & 8.357 & 5.071
			& 3.429 & 3.357 & 5.000 & \underline{2.357} & \textbf{2.143} \\
			\midrule
			\multirow{4}{*}{Length $L$}
			& $L\leq150$ ($n=31$)
			& 8.290 & 8.258 & 7.516 & 7.758 & 4.790
			& 3.823 & 4.532 & 4.613 & \underline{2.887} & \textbf{2.532} \\
			& $151\leq L\leq500$ ($n=22$)
			& 8.636 & 8.114 & 6.295 & 8.818 & 6.091
			& 3.295 & 4.250 & 4.545 & \underline{2.682} & \textbf{2.273} \\
			& $501\leq L\leq1000$ ($n=16$)
			& 7.000 & 7.875 & 6.406 & 7.656 & 5.938
			& 4.875 & \underline{3.406} & 5.125 & 3.656 & \textbf{3.062} \\
			& $L>1000$ ($n=17$)
			& 8.294 & 8.529 & 7.824 & 7.941 & 6.059
			& 3.529 & 3.324 & 4.588 & \underline{2.500} & \textbf{2.412} \\
			\midrule
			\multirow{6}{*}{Type $T$}
			& Biomedical ($n=10$)
			& 9.300 & 8.900 & 7.800 & 6.800 & 4.800
			& 4.100 & \textbf{3.000} & 4.300 & \textbf{3.000} & \textbf{3.000} \\
			& Industrial ($n=21$)
			& 7.857 & 7.690 & 6.690 & 8.047 & 5.476
			& 4.476 & 3.976 & 4.571 & \underline{3.357} & \textbf{2.857} \\
			& Motion ($n=17$)
			& 8.000 & 7.941 & 6.529 & 8.206 & 5.971
			& 4.029 & 3.441 & 5.647 & \underline{2.971} & \textbf{2.265} \\
			& ImageOutline ($n=20$)
			& 7.850 & 7.900 & 7.150 & 8.400 & 5.300
			& 3.675 & 4.450 & 4.800 & \underline{2.875} & \textbf{2.600} \\
			& Spectral ($n=11$)
			& 8.545 & 8.818 & 7.773 & 7.773 & 6.091
			& 2.955 & 4.455 & 4.136 & \underline{2.409} & \textbf{2.045} \\
			& Simulated ($n=7$)
			& 7.857 & 9.286 & 7.000 & 8.857 & 6.143
			& 2.786 & 5.000 & 3.786 & \textbf{2.071} & \underline{2.214} \\
			\bottomrule
	\end{tabular}}
\end{table*}

\subsubsection{Contribution of the Evolutionary Program Construction}
The first perspective demonstrates the contribution of the evolutionary program construction process by comparing EvoTSC against variants that bypass evolution entirely and rely solely on predefined feature extraction operators. Specifically, four variants are constructed by directly applying a single extraction operator (ShapeInc, ShapeDec, ShapePeak, or StatisDist) to the raw time series. The fifth variant, denoted AllFea, applies all four extraction operators and concatenates the resulting feature vectors into a single representation, providing an upper-bound baseline for non-evolutionary feature extraction. As shown in Table \ref{tab:ablation}, the four single-operator variants consistently occupy the lowest-ranked positions across all fourteen dataset groups. This  demonstrates that applying a single predefined extraction operator to the raw time series, without the evolutionary program construction, yields features of limited discriminative power. While AllFea consistently outperforms the individual operators by leveraging the complementarity among different extraction operators, it exhibits noticeable degradation relative to the full EvoTSC in every dataset group. This gap confirms that the performance gains of EvoTSC are not merely attributable to the availability of diverse extraction operators, but rather stem from the ability of the evolutionary search to discover task-specific program compositions tailored to the target dataset.

\subsubsection{Contribution of the Multi-layer Program Structure}
The second perspective investigates the contribution of the functional layers within the proposed multi-layer program structure through three degraded variants. The first variant, wo\_Dom, removes the domain transform layer to examine whether allowing the evolutionary search to exploit frequency-domain and difference-domain representations contributes to the performance. The second variant, wo\_Patch, removes the content-adaptive patching layer to assess whether segmenting the input into overlapping patches, rather than treating the full segment as a single extraction unit, enhances feature learning. The third variant, wo\_Con, removes the feature concatenation layer to examine whether allowing the evolution to simultaneously utilize features from multiple sources contributes to the performance. It can be seen from Table \ref{tab:ablation} that all three layer-removal variants exhibit degradation relative to the full EvoTSC across all the dataset groups, confirming that the functional layers contribute positively to the overall feature learning process.

\subsubsection{Effectiveness of the Pareto Tournament Selection}
The third perspective examines the effectiveness of the tailored Pareto tournament selection strategy. In this variant, denoted as wo\_Pareto, the tailored Pareto tournament selection is replaced with standard tournament selection, where the single averaged cross-validation accuracy serves as the selection criterion. As reported in Table \ref{tab:ablation}, this modification leads to performance degradation relative to the full \textsc{EvoTSC} in eleven out of fourteen dataset groups, with only two groups showing a tiny advantage for wo\_Pareto. This confirms that the proposed selection strategy contributes positively to the generalization performance of the learned features.

In summary, the ablation results demonstrate that each component of EvoTSC contributes meaningfully to the overall performance, collectively enabling the proposed approach to evolve effective and generalizable feature learning models for TSC. Furthermore, a fold-to-fold standard deviation comparison shows that the tailored Pareto tournament selection favors models performing consistently well across varying folds of the training data, and a sensitivity analysis indicates that the performance is insensitive to the choice of the tournament ratio $\mu$. Details of both analyses are provided in Section \Romannum{3} of the supplementary material.

\subsection{Analysis of the Evolved EvoTSC Individuals}
\label{s62}
\subsubsection{Structural Composition Analysis of Evolved EvoTSC Individuals}
The first experiment examines the structural composition of the evolved EvoTSC individuals across all datasets, aiming to answer a central question: what structural preferences does evolution discover across different categories of TSC problems, and what implicit domain knowledge do these preferences reveal?

\begin{figure*}[t]
	\centering
	\includegraphics[width=1\textwidth]{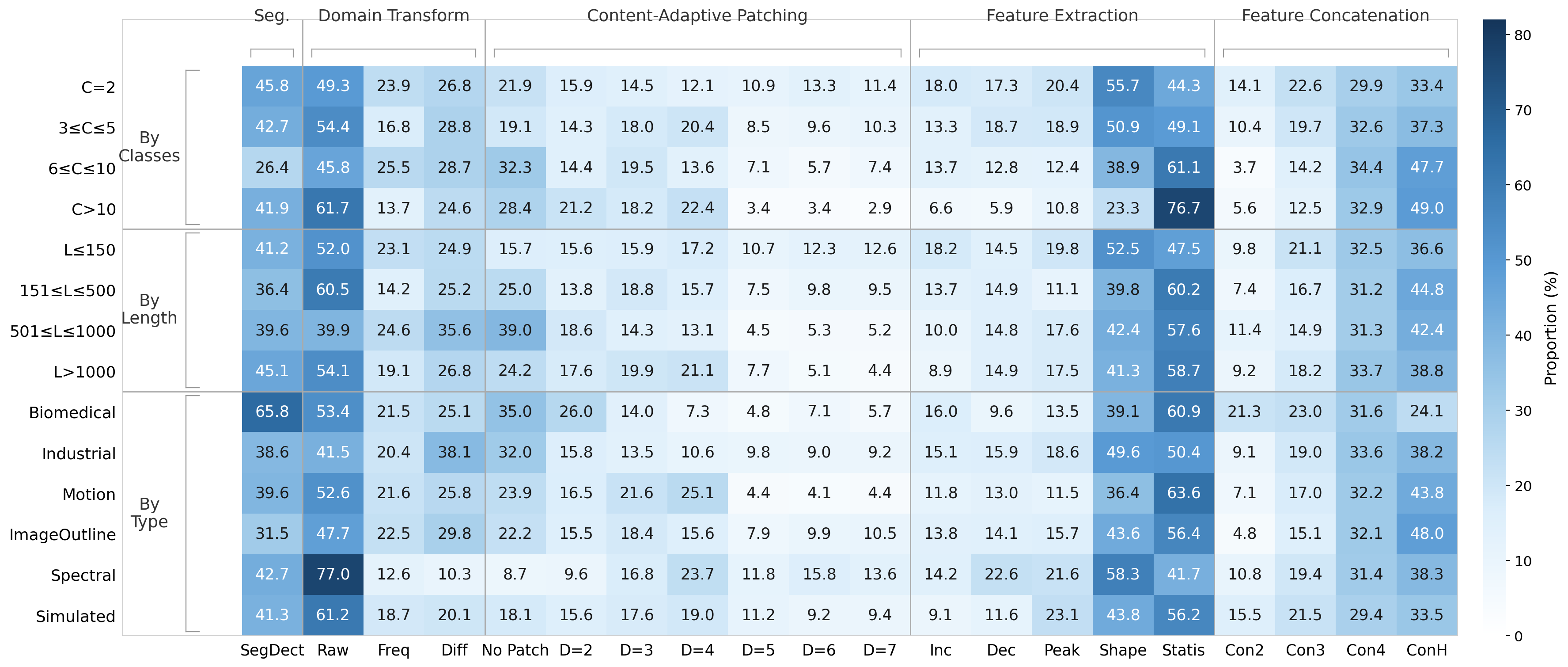}
	\vspace{-20pt}
	\caption{Heatmap of operation proportions across different dataset groups. Datasets are grouped by number of classes, time-series length, and data type. \textbf{Segment Detection (Seg.)}: \textit{SegDect} indicates the proportion of branches that use segment detection. \textbf{Domain Transform}: \textit{Raw} denotes the proportion of branches that operate directly in the original domain without applying any transformation; \textit{Freq} and \textit{Diff} denote frequency transform and differencing, respectively. \textbf{Content-Adaptive Patching}: $D=i$ indicates that the length of patches is $\lfloor \ell_{\text{in}}/2^{i-1} \rfloor$; \textit{No\,Patch} indicates that no patching operation is applied in the branch. \textbf{Feature Extraction}: values represent the proportions of different extraction operators, where \textit{Shape} aggregates all shape-related extraction operations.}
	\label{heatmap}
\end{figure*}

We collect the evolved EvoTSC individual from each of 30  runs for each of the 86 UCR datasets, yielding 2{,}580 program trees in total. Following the three grouping schemes introduced earlier, the operation usage within each dataset group is counted and the proportion of each operation is computed. For the feature concatenation layer operations, proportions are calculated relative to the total number of concatenation nodes in the corresponding group. For all other operations, proportions are calculated relative to the number of feature learning branches in that group. The following observations can be drawn from the heatmap in Fig. \ref{heatmap}.

For segment detection usage across application types, the \emph{Biomedical} group exhibits the highest proportion ($65.8\%$), far exceeding all other groups. This is consistent with clinical practice, where localizing specific waveform components, such as the QRS complex in ECG signals or characteristic deflections in EOG recordings, is essential for diagnosis. Across different series lengths, the segment detection rate rises from $36.4\%$ for the \emph{151\,$\leq$\,L\,$\leq$\,500} group to $45.1\%$ for the \emph{L\,$>$\,1000} group, consistent with the intuition that longer sequences are more likely to contain task-irrelevant or noisy regions, making explicit subsequence localization increasingly beneficial.

For domain transform usage across application types, the \emph{Industrial} group shows the highest proportion of difference domain usage ($38.1\%$), aligning with the nature of industrial monitoring signals, where abrupt state transitions and relative temporal variations often carry discriminative information. Conversely, the \emph{Spectral} group exhibits the highest raw domain usage ($77.0\%$). While this may appear counter-intuitive, the raw signals in these UCR datasets are already spectral curves, so the discriminative information inherently resides in the original domain. Applying a further frequency transform would amount to a redundant second-order transformation, and evolution therefore overwhelmingly favors operating directly in the original domain.

For the proportions of concatenation operators across different numbers of classes, the proportion of $\operatorname{\mathbf{FeaConH}}$ increases as the number of classes grows, from $33.4\%$ in the \emph{C\,=\,2} group to $49.0\%$ in the \emph{C\,$>$\,10} group. This is consistent with the intuition that more complex classification tasks with a larger number of classes require richer feature representations to construct sufficiently discriminative boundaries.

Overall, these results indicate that evolution is able to discover some structural preferences across different categories of TSC problems, and that the resulting preferences reveal meaningful implicit domain knowledge about how discriminative representations should be learned.
\begin{figure}[t]
	\centering
	\includegraphics[width=0.45\textwidth]{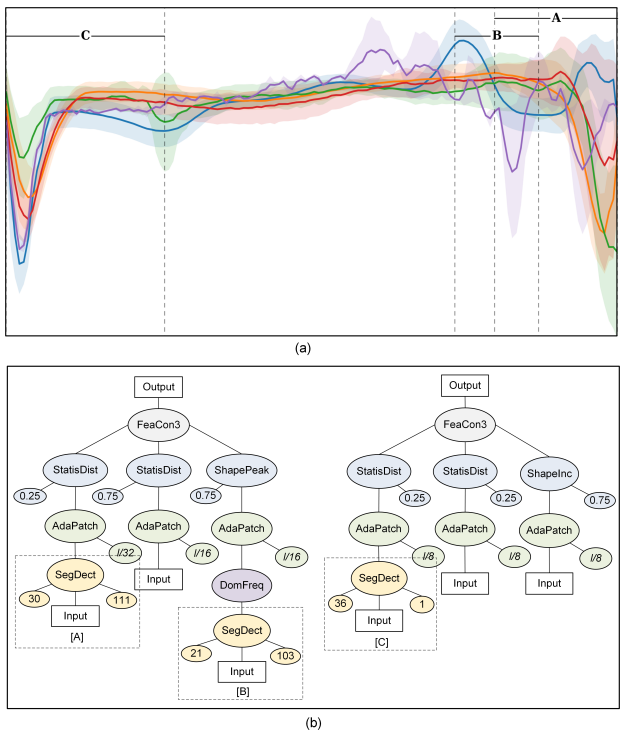}
	\vspace{-5pt}
	\caption{Case study on the ECG5000 dataset. (a) Visualization of the training instances, where different colors denote different classes. (b) Two evolved EvoTSC individuals, where each $\operatorname{\mathbf{SegDect}}$ node has a left terminal indicating the segment length and a right terminal indicating the starting position.}
	\label{casestudy}
\end{figure}

\subsubsection{An Example of the Evolved EvoTSC Individuals}
This section presents a case study to investigate whether the decision-making process of evolved EvoTSC individuals can be logically inspected and whether they can selectively focus on discriminative segments within a time series. ECG5000 is used as the example dataset, as ECG signals possess well-understood morphological components, which provide a clinical reference for verifying whether the detected segments are physiologically meaningful.

For each class, the per-time-step mean and standard deviation across all training instances are computed and visualized as solid curves with shaded bands in Fig. \ref{casestudy}(a). This representation is adopted because plotting all raw instances would be visually cluttered, while displaying only a subset could give a biased impression. Furthermore, two  evolved EvoTSC program trees from this dataset are illustrated in Fig. \ref{casestudy}(b), and the three segments they detect ([A],[B] and [C]) are highlighted on the corresponding regions in Fig. \ref{casestudy}(a).

The evolved EvoTSC individuals can be analyzed. For instance, analyzing the rightmost feature learning branch of the left individual in Fig. \ref{casestudy}(b), the original time series is first truncated to a segment starting at index 103 with a length of 21 (i.e., the interval $[103, 123]$). This segment is then transformed into the frequency domain via the $\operatorname{\mathbf{DomFreq}}$ operator, followed by a patching operation ($\operatorname{\mathbf{AdaPatch}}$) with a patch size of $\lfloor \ell_{\text{in}}/16 \rfloor$. Finally, the $\operatorname{\mathbf{ShapePeak}}$ operator is applied to extract features from the frequency representation. Such explicit, step-by-step reasoning allows domain practitioners to logically inspect the model's decision-making process.
	
Moreover, the segments highlighted by the evolved EvoTSC individuals exhibit clear inter-class separability. As illustrated in Fig. \ref{casestudy}(a), Segment [C] (positions $[1,36]$) covers the initial part of the heartbeat, where pronounced inter-class differences can be observed. This region contains a sharp deflection, suggesting that it corresponds to the clinically informative QRS complex in ECG interpretation. Segments [A] (positions $[111,140]$) and [B] (positions $[103,123]$) jointly span the terminal portion of the heartbeat, a region likely corresponding to the T-wave and the subsequent recovery phase. This region also exhibits greater inter-class separation than the middle portion of the series. These observations suggest that the evolutionary search can naturally identify and focus on physiologically meaningful regions.

Overall, this case study demonstrates that EvoTSC can evolve models with greater inspectability than black-box neural models. Additional studies on the Trace (industrial monitoring) and GunPoint (motion recognition) datasets are reported in the supplementary material and show similar patterns.

\subsection{Analytical Resource Estimates of the Evolved EvoTSC Models}
\label{s63}

This experiment analytically estimates the computational resource requirements of the evolved EvoTSC model and the baselines using two hardware-independent metrics: operation count, which reflects inference latency, and peak activation memory (in bytes), which reflects
memory consumption during inference \cite{lin2020mcunet}.
This analysis is conducted on the ECG5000 dataset (series length 140) under single-instance inference in float32 precision. ECG5000 is selected because ECG monitoring on wearable or implantable devices is an important resource-constrained application \cite{ECGTCN}. The operation count is estimated as the total number of floating-point arithmetic operations and comparisons across all layers or operators. Peak activation memory is estimated as the largest combined size of the input and output tensors of any layer or operator. Model storage is not included in this analysis, because model parameters are typically stored in the flash memory of microcontrollers rather than in SRAM, which is the focus of this analysis and is often the more constrained resource \cite{lin2020mcunet}. Additionally, these estimates cover only the feature learning stage and exclude downstream classifiers. These exclusions are  applied uniformly to all methods. 

\begin{figure}[ht]
	\centering
	\includegraphics[width=0.95\linewidth]{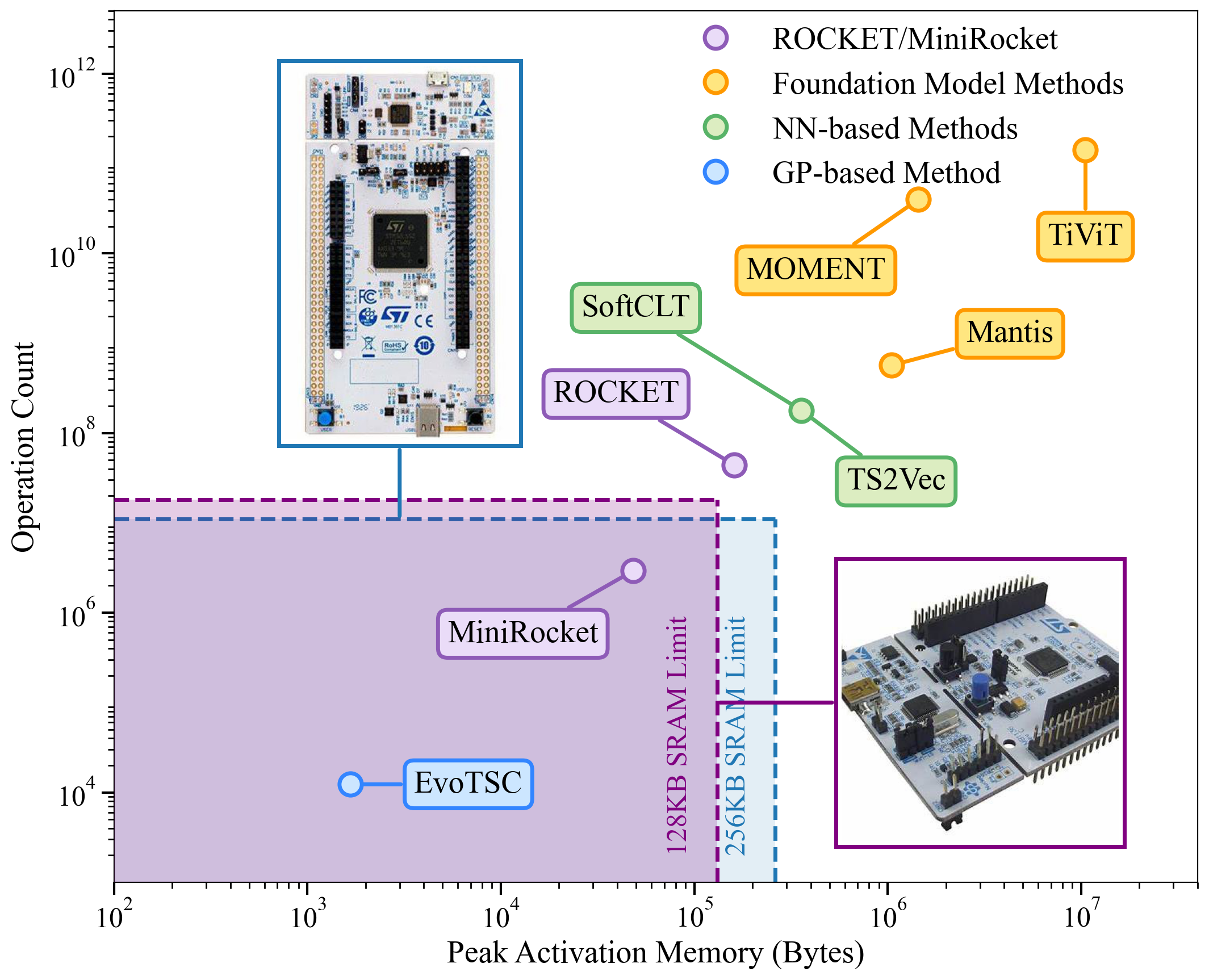}
	\vspace{-5pt} 
	\caption{Analytical resource estimates for the evolved EvoTSC model and baseline methods on the ECG5000 dataset (series length 140). The operation count is estimated as the total number of floating-point arithmetic operations and comparisons across all layers or operators. Peak activation memory is estimated as the largest combined size of the input and output tensors of any layer or operator. The vertical dashed lines indicate the SRAM limits of each microcontroller. The horizontal dashed lines represent idealized operation budgets for 100 ms of execution, assuming one operation per clock cycle. }
	\label{fig:resource}
\end{figure}

Fig. \ref{fig:resource} plots the estimated operation count against the peak activation memory on a log-log scale \cite{11030860} for the evolved EvoTSC model and seven baseline models. To make these values more intuitive, we map them against the hardware constraints of two widely used commercial microcontrollers: the STM32F446RE \footnote{https://www.st.com/en/microcontrollers-microprocessors/stm32f446re.html} (operating at up to 180 MHz with 128 KB SRAM) and the STM32L552ZE \footnote{https://www.st.com/en/microcontrollers-microprocessors/stm32l552ze.html} (operating at up to 110 MHz with 256 KB SRAM) \cite{king2025micronas}. The vertical dashed lines indicate the SRAM limits of each microcontroller. The horizontal dashed lines represent idealized operation budgets for 100 ms of execution, assuming one operation per clock cycle. 

As shown in Fig. \ref{fig:resource}, EvoTSC occupies the lower-left region of the plot, requiring approximately $10^{4}$ operations and on the order of $10^{3}$ bytes of peak activation memory. Both values lie well below the reference budgets of even the more constrained STM32F446RE microcontroller. Among the baseline methods, only MiniRocket also lies within the reference budgets of both devices, but it still requires about two orders of magnitude more operations and an order of magnitude more memory than the model evolved by EvoTSC. The NN-based methods, whose two points nearly coincide because SoftCLT and TS2Vec share the same network structure, require over $10^{8}$ operations and more than $10^{5}$ bytes of peak activation memory, surpassing both the SRAM capacity and the operation budget of the two microcontrollers. The foundation model methods exhibit even more prohibitive computational cost, with MOMENT and TiViT reaching approximately $10^{10}$ and $10^{11}$ operations respectively, and TiViT requiring around $10^{7}$ bytes of peak activation memory. This substantial reduction in computational cost stems from the fact that, rather than relying on large parameter matrices and dense tensor multiplications, the model evolved by EvoTSC is presented as a concise symbolic program. These results suggest that EvoTSC is a promising option for resource-constrained environments where deep NN methods are less practical.

Note that this analysis has two limitations. First, the operation budgets are based on the idealized assumption of one operation per clock cycle. Second, inference latency has not been measured on actual hardware. Therefore, these estimates are intended only for analytical resource comparison rather than as evidence of deployment feasibility.

\section{Conclusions and Future Work}
\label{s7}
This paper proposes EvoTSC, a GP-based approach that automatically evolves lightweight and effective feature learning models for TSC. The key insight behind EvoTSC is that by embedding diverse forms of prior expert knowledge into a carefully designed multi-layer program structure, the evolutionary search can be guided to discover effective compositions of TSC-specific operations tailored to the target dataset. Additionally, a tailored Pareto tournament selection strategy is introduced to enhance the generalizability of the learned features by favoring models that generalize well across varying data subsets during the evolutionary search. Building upon the evolved EvoTSC models, an ensemble classifier called EvoTSC-E is further developed for scenarios where computational resources are less constrained.

Experimental results yielded several important findings. First, compared with each feature-based method, the evolved feature learning models either perform significantly better or show no significant difference on about 70\% of the datasets, while EvoTSC-E achieves the best average rank among all the 17 compared methods. Second, ablation studies confirm that each proposed component contributes meaningfully to the overall performance. Finally, further analyses show that EvoTSC evolves highly inspectable models with computational costs that are orders of magnitude lower than those of NN-based methods.

Despite the promising results, several directions remain open for future investigation. First, the current work focuses exclusively on univariate time series. Extending EvoTSC to handle multivariate time series, where cross-variate dependencies carry additional information, is a natural and important direction. This extension may require redesigning the program structure to incorporate cross-variate operators that can capture correlations among multiple variates. Second, this work considers the offline setting, where each instance is a complete series that is classified as a whole. Extending EvoTSC to streaming scenarios, where data arrive incrementally and decisions may be required before the full series is observed, is an interesting research direction, which requires re-design of the program structure as well as the fitness function.

\bibliographystyle{IEEEtran}
\bibliography{refs}

\end{document}